\documentclass{article}

\usepackage{microtype}
\usepackage{graphicx}
\usepackage{subcaption}
\usepackage{booktabs} 

\usepackage{hyperref}

\usepackage[accepted]{icml2024}

\usepackage{amsmath}
\usepackage{amssymb}
\usepackage{mathtools}
\usepackage{amsthm}

\usepackage[capitalize,noabbrev]{cleveref}
\usepackage{xcolor}
\definecolor{forestgreen}{rgb}{0.333, 0.745, 0.333}

\theoremstyle{plain}

\theoremstyle{definition}

\theoremstyle{remark}

\usepackage[textsize=tiny]{todonotes}

\def\etal{\emph{et al.}}

\def\ie{\emph{i.e.}}

\DeclareUnicodeCharacter{2212}{-}
\newlength\savewidth
\usepackage{makecell}
\usepackage{multirow}
\usepackage{pifont}

\icmltitlerunning{Vision Superalignment: Weak-to-Strong Generalization for Vision Foundation Models}

\begin{document}

\twocolumn[
\icmltitle{Vision Superalignment: Weak-to-Strong Generalization \\ for Vision Foundation Models}



\icmlsetsymbol{equal}{*}

\begin{icmlauthorlist}
\icmlauthor{Jianyuan Guo}{equal,comp,usyd}
\icmlauthor{Hanting Chen}{equal,comp}
\icmlauthor{Chengcheng Wang}{comp}
\icmlauthor{Kai Han}{comp}
\icmlauthor{Chang Xu}{usyd}
\icmlauthor{Yunhe Wang}{comp}
\end{icmlauthorlist}

\begin{center}
jianyuan\_guo@outlook.com; \{chenhanting,wangchengcheng11,kai.han\}@huawei.com; c.xu@sydney.edu.cn;
\end{center}

\icmlaffiliation{usyd}{School of Computer Science, Faculty of Engineering, University of Sydney, Sydney, Australia}
\icmlaffiliation{comp}{Huawei Noah's Ark Lab, Beijing, China}

\icmlcorrespondingauthor{Yunhe Wang}{yunhe.wang@huawei.com}

\icmlkeywords{Machine Learning, ICML}

\vskip 0.3in
]



\printAffiliationsAndNotice{\icmlEqualContribution} 

\begin{abstract}
Recent advancements in large language models have sparked interest in their extraordinary and near-superhuman capabilities, leading researchers to explore methods for evaluating and optimizing these abilities, which is called superalignment. In this context, our paper delves into the realm of vision foundation models, focusing on the concept of weak-to-strong generalization, which involves using a weaker model to supervise a stronger one, aiming to enhance the latter's capabilities beyond the former's limits. We introduce a novel and adaptively adjustable loss function for weak-to-strong supervision. Our comprehensive experiments span various scenarios, including few-shot learning, transfer learning, noisy label learning, and common knowledge distillation settings. The results are striking: our approach not only exceeds the performance benchmarks set by strong-to-strong generalization but also surpasses the outcomes of fine-tuning strong models with whole datasets. This compelling evidence underscores the significant potential of weak-to-strong generalization, showcasing its capability to substantially elevate the performance of vision foundation models. The code is available at \href{https://github.com/ggjy/vision_weak_to_strong}{https://github.com/ggjy/vision\_weak\_to\_strong}.
\end{abstract}

\section{Introduction}

The evolution and maturation of artificial intelligence are profoundly reliant on human evaluation, guidance, and experience. In the realm of computer vision, convolutional networks acquire semantic knowledge of images through extensive labeling provided by experts, such as object boundaries in the COCO dataset~\cite{lin2014microsoft} or image categories in ImageNet~\cite{deng2009imagenet}. Similarly, in fields like robotics, reinforcement learning~\cite{kaelbling1996reinforcement} often depends on human-defined reward functions to steer machines towards optimal performance. In the domain of Natural Language Processing (NLP), recurrent neural networks~\cite{hochreiter1997long} and Transformers~\cite{vaswani2017attention} are capable of learning the distribution of language from vast amounts of unsupervised text generated by humans. This synergy suggests that AI models are essentially advancing while standing on the shoulders of human intelligence, leveraging the depth and breadth of human expertise to reach new heights of capability and understanding.

\begin{figure}
\centering
\includegraphics[width=1\linewidth]{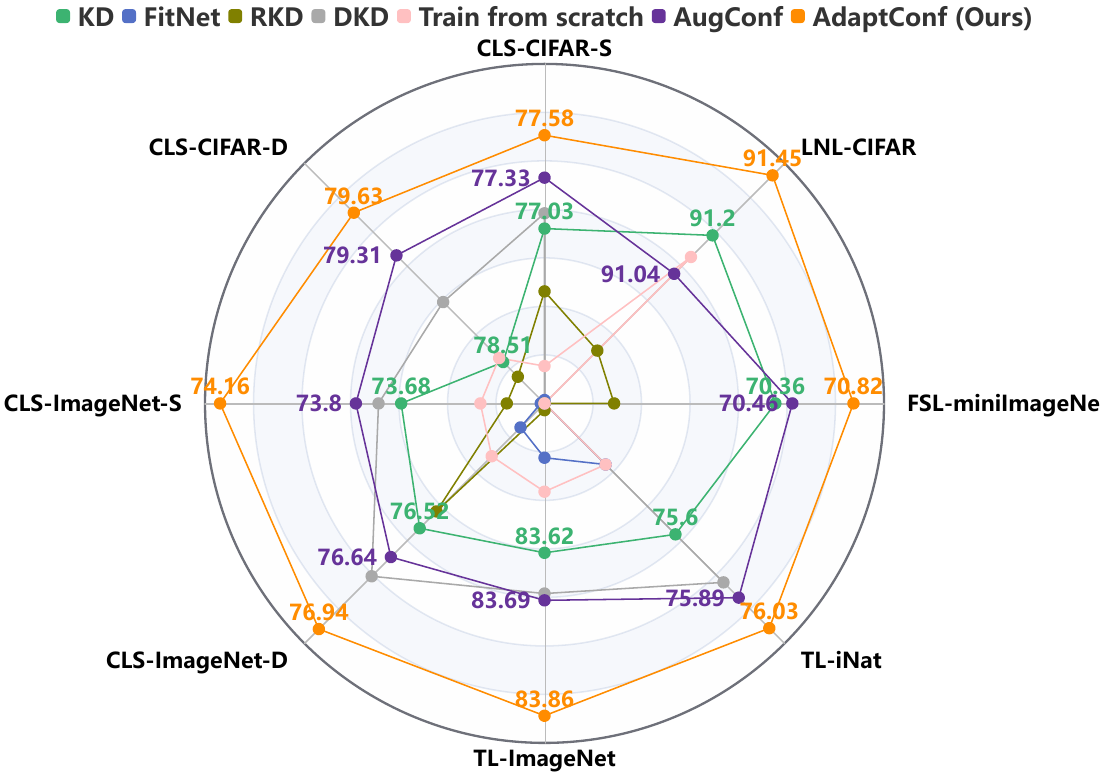}
\vspace{-10pt}
\caption{\small{Our proposed AdaptConf achieves the best performance on a broad range of tasks compared with other knowledge distillation based methods. The corresponding values are calculated by averaging results on eack task. CLS-CIFAR-S: Table~\ref{tab:cifar_cls_same}, CLS-CIFAR-D: Table~\ref{tab:cifar_cls_different_gt}, CLS-ImageNet-S: Table~\ref{tab:imagenet_cls}, CLS-ImageNet-D: Table~\ref{tab:imagenet_cls}, TL-ImageNet: Table~\ref{tab:imagenet_transfer}, TL-iNat: Table~\ref{tab:inat_transfer}, FSL-miniImageNet: Table~\ref{tab:meta_base-class}, LNL-CIFAR: Table~\ref{tab:lnl}.}}
\vspace{-2pt}
\label{fig:hexagon}
\end{figure}

The landscape of deep learning has undergone a transformative shift, with neural networks increasingly demonstrating capabilities that surpass human performance across various domains. For instance, AlphaGO~\cite{silver2016mastering} showcased an ability in the game of Go that far exceeded the prowess of the strongest human players. Similarly, GPT-like models~\cite{brown2020language} are capable of achieving results beyond the average human level in a variety of exams. Notably, this phenomenon emerged even earlier in the field of computer vision. As early as 2015, meticulously designed convolutional neural networks~\cite{he2015delving} were already achieving results on large-scale visual classification tasks like ImageNet that surpassed human performance. This trend of superhuman achievements has driven the research community to focus on how to control, evaluate, and optimize these exceptionally capable models, recognizing the immense potential they hold for advancing our understanding and application of artificial intelligence.

To address the intricate challenge of leveraging human expertise in supervising superhuman AI models, the concept of ``superalignment" has been introduced. This approach aims to align superhuman models in a way that maximizes their learning from human input. A seminal work in this area is the concept of Weak-to-Strong Generalization (WSG)~\cite{burns2023weak}. This research presents an intriguing analogy to explore the feasibility of using weaker models to supervise stronger ones. The results from this line of inquiry are stimulating: strong models, already endowed with robust generalization and representational capabilities, can achieve performances that surpass their weaker counterparts through simple supervision provided by these weaker models. This supervision often involves incomplete or flawed labels, yet the stronger models can effectively transcend these limitations. Such findings have not only affirmed the viability of Weak-to-Strong Generalization but have also demonstrated its efficacy in fields like natural language processing and reinforcement learning.

In this paper, we delve into the topic of ``vision superalignment," specifically investigating the applicability of Weak-to-Strong Generalization (WSG) within the context of vision foundation models. Our study meticulously designs and examines multiple scenarios in computer vision, including few-shot learning, transfer learning, noisy label learning, and traditional knowledge distillation settings. In these scenarios, stronger models are trained to learn from weaker models. Through detailed validation and comparative experiments, we demonstrate the feasibility of WSG in the visual domain. Furthermore, we introduce an improved and adaptive confidence scheme to enhance the efficacy of WSG. Our study not only validates the concept of WSG in vision but also contributes significantly to the broader pursuit of superalignment across various AI modalities. Our work represents a substantial step forward in understanding and optimizing the interaction between human-level expertise and superhuman AI capabilities, potentially paving the way for groundbreaking advancements in artificial intelligence.

\section{Related Works}
The pursuit of enhancing the performance of deep neural networks in computer vision has led to the development of the teacher-student learning paradigm~\cite{kd}. This approach typically involves a stronger model (teacher) improving the performance of a weaker model (student), with extensive research focusing on optimizing the capabilities of the weaker model. Various strategies have been proposed to achieve this. For instance, ~\cite{hint} suggests that in addition to the output logits, incorporating intermediate layer features for supervision can significantly boost the student's learning. ~\cite{rkd} posits that the relationships between samples can serve as valuable supervisory information.

In a further refinement of this approach, ~\cite{dkd} redefines classical knowledge distillation (KD) loss, segmenting it into target-class and non-target-class distillation to balance the transfer of these two types of information more effectively. ~\cite{ofd} delves into the details and components of feature distillation, arriving at an improved method for the transfer of feature knowledge. Meanwhile, ~\cite{revkd} explores cross-stage feature transfer as an alternative to the conventional same-stage feature transfer. These methods have proven effective for strong-to-weak generalization scenarios.

However, with the gradual increase in the size and complexity of vision foundation models, the focus has shifted towards weak-to-strong generalization, \ie, how a weak model can improve a strong model. In this context, \cite{furlanello2018born} investigates knowledge distillation between teachers and students of equal size, demonstrating the feasibility of distilling models of the same size. Building upon this, \cite{xie2020self} introduces the use of additional unlabeled data for knowledge distillation among models of equal size, further validating the effectiveness of strong-to-strong generalization, especially in scenarios with abundant data availability. This body of research sets the stage for our exploration into weak-to-strong generalization, a relatively uncharted yet promising domain in the field of vision foundation models.

\section{Vision Superalignment}

In order to investigate how to supervise and optimize superhuman computer vision models, our focus centers on the study of weak-to-strong generalization for vision foundation models. In this section, we initially delve into examining and defining what constitutes vision foundation models. These models, characterized by their extensive capacity and versatility, form the backbone of our research. Subsequently, we address a critical challenge inherent in the weak-to-strong generalization approach: the inaccuracy of output labels from weak models. In response to this, we introduce an innovative solution – adaptive confidence distillation. This method is designed to enhance the learning process of strong models by effectively utilizing the guidance provided by weak models. Adaptive confidence distillation operates on the principle that even imperfect or partially accurate guidance from a weak model can be a valuable learning tool for a stronger model, provided that the information is processed and adapted correctly.

\subsection{Vision Foundation Models}

In our exploration of weak-to-strong generalization for vision foundation models, it is crucial to first define what constitutes these models. There are several potential categories of candidates that represent vision foundation models, each characterized by their unique capabilities and approaches in the realm of computer vision.

\textbf{Text-Visual Fusion Models:} The first category includes models that integrate visual and linguistic tasks. A notable example is the work of Radford et al.~\cite{radford2021learning}, which constructs a foundational model by aligning computer vision tasks with language tasks through image-text pre-training pairs. This approach bridges the gap between textual and visual information, providing a comprehensive understanding of both domains.

\textbf{Image Generation Models:} The second category focuses on models that are capable of generating images, which can be considered as a form of modeling the image space. Rombach et al.\cite{rombach2022high} demonstrate this through their ability to generate a plethora of images from textual descriptions, establishing a basis for image generation models. Similarly, Chen et al.\cite{chen2020generative} employ a GPT-like pre-training method for images, resulting in a generative Transformer model with significant generalization capabilities in image creation.

\textbf{Architecture for General or Zero-Shot Visual Tasks:} The third category seeks to develop architectures capable of solving a range of visual tasks, either generally or in a zero-shot manner. Bai et al.\cite{bai2023sequential} approach this by modeling a series of image tasks as sequential challenges, creating a large vision model that addresses a spectrum of visual problems. Additionally, Kirillov et al.\cite{kirillov2023segment} propose the ``Segment Anything" model, achieving impressive zero-shot segmentation results.

In our quest to identify the most suitable candidates for vision foundation models to be used in weak-to-strong generation tasks, we propose a definition focused on versatility and effectiveness in the visual domain. We posit that vision foundation models should be applicable to a broad range of visual tasks while delivering high-quality performance.

Based on this criterion, we suggest that backbones pretrained on ImageNet represent strong contenders as vision foundation models. The rationale for this choice is twofold. Firstly, these backbones have proven to be highly adaptable and effective for key tasks in computer vision, such as classification, detection, and segmentation. By fine-tuning these backbones, state-of-the-art (SOTA) accuracies can be achieved in these tasks, demonstrating their robustness and versatility. Secondly, there is an extensive array of pretraining algorithms developed specifically for these models~\cite{he2022masked}, which further qualifies them as universal pretraining models for vision tasks. Additionally, these types of models are often used as one of the branches in vision-language multimodal models~\cite{du2022survey}, highlighting their applicability in cross-modal tasks.

Therefore, for our experimental analysis, we choose to focus on these backbone models as representatives of vision foundation models. We aim to conduct our weak-to-strong generalization analysis using the fundamental task of image classification as a baseline. This approach allows us to thoroughly assess the capabilities and potential of weak-to-strong generation in a controlled yet comprehensive manner, offering insights that could be extrapolated to other, more complex vision tasks.

\begin{table*}[h]
\renewcommand{\arraystretch}{1.1}
\setlength{\belowcaptionskip}{-10pt}
\center
\begin{small}
\begin{tabular}{c|cccccc}
\multirow{2}{*}{Teacher} & ResNet20 & ResNet32 & ResNet8$\times$4 & WRN-16-2 & WRN-40-1 & VGG8 \\
& 68.93 & 71.72 & 72.41 & 72.71 & 72.30 & 71.99 \\
\multirow{2}{*}{Student} & ResNet56 & ResNet110 & ResNet32$\times$4  & WRN-40-2 & WRN-40-2 & VGG13 \\
& 72.94 & 74.80 & 79.90 & 77.20 & 77.20 & 75.26 \\ \Xhline{3\arrayrulewidth} 
KD~\cite{kd} & 73.81 & 76.45 & 79.32 & 78.25 & 77.97 & 76.41 \\
FitNet~\cite{hint} & 70.51 & 73.15 & 77.65 & 76.71 & 76.12 & 76.39 \\
RKD~\cite{rkd} & 72.98 & 75.62 & 80.10 & 77.27 & 77.76 & 76.20 \\
ReviewKD~\cite{revkd} & 70.15 & 72.30 & 77.22 & 75.86 & 75.78 & 74.22 \\
DKD~\cite{dkd} & \underline{73.90} & 76.57 & 79.52 & 78.18 & 77.95 & \underline{76.62} \\
AugConf~\cite{burns2023weak} & 73.86 & \underline{76.72} & \underline{80.34} & \underline{78.34} & \underline{78.15} & 76.55 \\
\hline
AdaptConf~(\textbf{Ours}) & \textbf{74.17} & \textbf{76.86} & \textbf{80.64} & \textbf{78.58} & \textbf{78.40} & \textbf{76.84} \\
$\Delta$ & \textcolor{forestgreen}{+1.23} & \textcolor{forestgreen}{+2.06} & \textcolor{forestgreen}{+0.74} & \textcolor{forestgreen}{+1.38} & \textcolor{forestgreen}{+1.20} & \textcolor{forestgreen}{+1.58} \\
\end{tabular}
\end{small}
\vspace{-2pt}
\caption{\textbf{Results on the CIFAR-100 validation set.} Teachers and students are in the \textbf{same} architectures. And $\Delta$ represents the performance improvement over the student model trained from scratch. All results are the average over 3 trials.}
\label{tab:cifar_cls_same}
\end{table*}

\subsection{Adaptive Confidence Distillation}

In this subsection, we explore the methodology for implementing weak-to-strong generalization in vision foundation models. The central question we address is how a weak vision foundation model can supervise a stronger counterpart effectively. ~\cite{burns2023weak} proposes an augmented confidence loss approach, which is formulated as:
\begin{equation}
L_{\mbox{\small conf}}(f) = (1 − \alpha)  \mbox{CE}(f(x), f_w(x)) + \alpha \mbox{CE}(f(x), \hat{f}(x)),
\label{eq:augconf}
\end{equation}
where $f$ represent the strong model that needs to be optimized, and  $f_w$ denote the weak model, $\hat{f}(x)$ refers to the hard label predicted by the strong model for an input image $x$. The loss function incorporates the cross-entropy loss (CE) and is balanced by a hyperparameter $\alpha$. In this formulation, the first term of the loss function resembles the traditional knowledge distillation loss, signifying the learning process of the strong model from the weak model. Given that the labels provided by the weak model may not always be accurate, the second term of the loss function encourages the strong model to leverage its superior generalization abilities and prior knowledge to refine its predictions.

The strength of this approach lies in its ability to balance direct learning from the weak model with the strong model's intrinsic capacity for understanding and interpreting the visual data. This method paves the way for the strong model to surpass the limitations of the weak model, utilizing the latter's guidance while simultaneously enhancing its predictions through its advanced capabilities.

Addressing the limitations inherent in the supervision provided by weak models and the inaccuracies of strong models' self-generated hard labels, a more sophisticated approach is required beyond a simple weighted combination of these labels. Given the challenge in directly discerning the accuracy of each label, leveraging confidence as a metric for selecting the most probable correct label emerges as a viable solution.

We propose to use the discrepancy between the soft label and the hard label as an indicator of the model's confidence. The underlying rationale is that when a model's soft label closely aligns with its hard label, it suggests a higher confidence in its own judgment. To capitalize on this insight, we introduce an adaptive confidence loss that dynamically adjusts based on the model's confidence level. The specific formulation of this loss is as follows:
\begin{equation}
\small
\begin{aligned}
&L_{\mbox{\scriptsize AC}}(f) = (1 − \beta(x))  \mbox{CE}(f(x), f_w(x)) + \beta(x) \mbox{CE}(f(x), \hat{f}(x)),\\
&\beta (x)= \frac{\mbox{exp}(\mbox{CE}(f(x), \hat{f}(x)))}{\mbox{exp}(\mbox{CE}(f(x), \hat{f}(x)))+\mbox{exp}(\mbox{CE}(f(x), \hat{f_w}(x)))}.
\label{eq:adaptconf}
\end{aligned}
\end{equation}
In this formula, $\beta(x)$ is a function of the input image $x$ that calculates the confidence weight and $\hat{f_w}(x)$ is the hard label of $x$ in the weak model. This weight determines the balance between learning from the weak model and relying on the strong model's own predictions. The cross-entropy loss (CE) is used for both components, with the first term focusing on learning from the weak model and the second term emphasizing the strong model's self-supervision.

This adaptive confidence loss enables a more nuanced approach to weak-to-strong generalization. By adjusting the weight based on confidence levels, it allows the strong model to discern when to prioritize its own predictions over the guidance of the weak model and vice versa. This adaptability is key to overcoming the inaccuracies and limitations of both models, leading to more effective learning and enhanced performance in vision foundation models.

\section{Experiment}
In this section, we report our main empirical results on various tasks, including baselines and promising methods. All implementation details are attached in supplementary materials.

\begin{table*}[t]
{
\begin{subfigure}{1.0\linewidth}
\small
\renewcommand{\arraystretch}{1.1}
\center
\begin{tabular}{c|ccccc}
\multirow{2}{*}{Teacher} & ShuffleNet-V1 & ShuffleNet-V1 & MobileNet-V2 & MobileNet-V2 & ShuffleNet-V2 \\
& 72.40 & 72.40 & 66.85 & 66.85 & 74.44 \\
\multirow{2}{*}{Student} & ResNet32$\times$4 & WRN-40-2 & VGG13 & ResNet50 & ResNet32$\times$4 \\
& 79.90 & 77.20 & 75.26 & 80.43 & 79.90 \\
\Xhline{3\arrayrulewidth}
KD~\cite{kd} & 80.19 & 78.02 & 75.39 & 78.64 & 80.31 \\
FitNet~\cite{hint} & 77.61 & 75.15 & 72.36 & 75.92 & 78.05 \\
RKD~\cite{rkd} & 80.30 & 77.23 & 76.21 & 79.89 & 80.39 \\
ReviewKD~\cite{revkd} & 78.43 & 75.98 & 73.69 & 77.05 & 77.84 \\
DKD~\cite{dkd} & 80.55 & \underline{78.10} & 75.81 & 79.65 & 80.67\\
AugConf~\cite{burns2023weak} & \underline{80.62} & 77.92 & \underline{76.43} & \underline{80.75} & \underline{80.84} \\
\hline
AdaptConf~(\textbf{Ours}) & \textbf{80.99} & \textbf{78.55} & \textbf{76.58} & \textbf{80.98} & \textbf{81.06} \\
$\Delta$ & \textcolor{forestgreen}{+1.09} & \textcolor{forestgreen}{+1.35} & \textcolor{forestgreen}{+1.32} & \textcolor{forestgreen}{+0.55} & \textcolor{forestgreen}{+1.16} \\
\end{tabular}
\vspace{-4pt}
\caption{Trained with teacher's prediction and ground truth label. $\Delta$ represents the improvement over the student trained from scratch.}
\label{tab:cifar_cls_different_gt}
\end{subfigure}
}
\\[-2pt]
{
\begin{subfigure}{1.0\linewidth}
\small
\renewcommand{\arraystretch}{1.1}
\center
\begin{small}
\begin{tabular}{c|ccccc}
\multirow{2}{*}{Teacher} & ShuffleNet-V1 & ShuffleNet-V1 & MobileNet-V2 & MobileNet-V2 & ShuffleNet-V2 \\
& 72.40 & 72.40 & 66.85 & 66.85 & 74.44 \\
Student & ResNet32$\times$4 & WRN-40-2 & VGG13 & ResNet50 & ResNet32$\times$4 \\
\Xhline{3\arrayrulewidth}
KD~\cite{kd} & 77.92 & \underline{76.45} & 72.13 & 73.32 & 78.27 \\
FitNet~\cite{hint} & 75.74 & 74.03 & 70.57 & 71.45 & 76.42 \\
RKD~\cite{rkd} & 76.59 & 75.70 & 70.28 & 72.06 & 77.84 \\
AugConf~\cite{burns2023weak} & \underline{78.25} & 76.37 & \underline{72.51} & \underline{74.48} & \underline{78.81} \\
\hline
AdaptConf~(\textbf{Ours}) & \textbf{78.48} & \textbf{76.66} & \textbf{72.93} & \textbf{74.67} & \textbf{79.04} \\
$\Delta$ & \textcolor{forestgreen}{+6.08} & \textcolor{forestgreen}{+4.26} & \textcolor{forestgreen}{+6.08} & \textcolor{forestgreen}{+7.82} & \textcolor{forestgreen}{+4.37} \\
\end{tabular}
\end{small}
\vspace{-1pt}
\caption{Trained with teacher's prediction only. $\Delta$ represents the performance improvement over the teacher model.}
\label{tab:cifar_cls_different_no_gt}
\end{subfigure}
}
\vspace{-6pt}
\caption{\textbf{Results on the CIFAR-100 validation set.} Teachers and students are in the \textbf{different} architectures. All results are the average over 3 trials.}
\label{tab:cifar_cls_different}
\end{table*}

\begin{table}[t]
{
\begin{subfigure}{1.0\linewidth}
\small
\renewcommand{\arraystretch}{1.05}
\setlength{\tabcolsep}{5pt}
\setlength{\belowcaptionskip}{-10pt}
\center
\begin{tabular}{c|c|c}
Teacher: ResNet50 (80.36) & Teacher + GT & Teacher \\
Student: ViT-B (MAE pretrain) & 83.53 & - \\
\Xhline{3\arrayrulewidth} 
KD~\cite{kd} & 83.62 & 82.32 \\
FitNet~\cite{hint} & 82.48 & 81.02 \\
RKD~\cite{rkd} & 82.19 & 80.98 \\
DKD~\cite{dkd} & 83.68 & - \\
AugConf~\cite{burns2023weak} & \underline{83.70} & \underline{82.38} \\
\hline
AdaptConf~(\textbf{Ours}) & \textbf{83.86} & \textbf{82.51} \\
$\Delta$ & \textcolor{forestgreen}{+0.33} & \textcolor{forestgreen}{+2.15} \\
\end{tabular}
\vspace{-4pt}
\caption{{Top-1 results on the ImageNet validation set.}}
\label{tab:imagenet_transfer}
\end{subfigure}
}
\\[2pt]
{
\begin{subfigure}{1.0\linewidth}
\small
\renewcommand{\arraystretch}{1.05}
\setlength{\tabcolsep}{5pt}
\setlength{\belowcaptionskip}{-10pt}
\center
\begin{tabular}{c|c|c}
Teacher: ResNet101 (67.42) & Teacher + GT & Teacher \\
Student: ViT-B (MAE pretrain) & 75.28 & - \\
\Xhline{3\arrayrulewidth} 
KD~\cite{kd} & 75.60 & 71.57 \\
FitNet~\cite{hint} & 73.68 & 70.11 \\
DKD~\cite{dkd} & 75.82 & - \\
AugConf~\cite{burns2023weak} & \underline{75.90} & \underline{71.73} \\
\hline
AdaptConf~(\textbf{Ours}) & \textbf{76.03} & \textbf{71.99} \\
$\Delta$ & \textcolor{forestgreen}{+0.75} & \textcolor{forestgreen}{+4.57} \\
\end{tabular}
\vspace{-4pt}
\caption{{Top-1 results on the iNaturalist 2019 test set.}}
\label{tab:inat_transfer}
\end{subfigure}
}
\vspace{-4pt}
\caption{\textbf{Transfer learning results.} The student model is a ViT-B~\cite{vit} pretrained by the self-supervised MAE framework~\cite{mae}. $\Delta$ denotes the performance improvement over student/teacher in second/third columns.}
\label{tab:transfer}
\end{table}

\subsection{Tasks}
\textbf{Image Classification.} 
Our experiments are primarily focused on two benchmark datasets. CIFAR-100~\cite{cifar} is a widely recognized dataset for image classification, comprising 32×32 pixel images across 100 categories, with training and validation sets containing 50,000 and 10,000 images, respectively. Conversely, ImageNet~\cite{deng2009imagenet} is a large-scale dataset for classification tasks, encompassing 1.28 million training images and 50,000 validation images across 1,000 classes. Additionally, we explore scenarios where only soft labels generated by a weak teacher model are available for training.

\textbf{Few-shot learning.} 
We explore few-shot learning across the miniImageNet~\cite{vinyals2016matching} dataset which contains 100 classes sampled from ILSVRC-2012~\cite{russakovsky2015imagenet}. We randomly split the dataset into 64, 16, and 20 classes as training, validation, and testing sets, respectively. And ensure that each class has 600 images of 84$\times$84 image size. We utilize the ResNet36 to explore the weak-to-strong generalization performance in few-shot task. To demonstrate weak-to-strong generalization performance, we follow Meta-Baseline and conduct related experiments on classifier stage and meta stage.

\textbf{Transfer learning.}
We explore transfer learning across two benchmark datasets: ImageNet~\cite{deng2009imagenet}, and iNaturalist 2018~\cite{inat}, the latter comprising 437,513 training images and 24,426 test images distributed across 8,142 species. We utilize the ViT-B~\cite{vit} that has been pretrained on the ImageNet training set using the self-supervised MAE~\cite{mae} approach, leveraging only image data without labels. Our results are reported for the fine-tuning phase, which is conducted under the guidance of a weak teacher model on each benchmark. Furthermore, we investigate scenarios in which only soft labels produced by the weak teacher model are used for training.

\textbf{Learning with noisy labels.} 
We evaluate our approach using two datasets with simulated label noise, specifically CIFAR-10~\cite{cifar} and CIFAR-100~\cite{cifar}. Consistent with prior research~\cite{li2020dividemix,tanaka2018joint}, we introduce two distinct types of simulated noisy labels: symmetric and asymmetric. Symmetric noise is introduced by randomly substituting the labels of a certain proportion of the training data with other possible labels uniformly. In contrast, asymmetric noise involves systematic mislabeling to mimic real-world errors, such as flipping the labels to closely related classes. For example, in CIFAR-10, \emph{truck} is mislabeled as \emph{automobile}, \emph{bird} as \emph{airplane}, and \emph{cat} is interchanged with \emph{dog}. For CIFAR-100, similar mislabeling is applied within each of the super-classes in a circular fashion.

\textbf{Baseline methods.}
The predominant framework for implementing teacher-student training paradigms is knowledge distillation~\cite{kd}. This approach outlines a method where a larger, more complex teacher network guides the training of a more compact student network. Nonetheless, inspired by the findings of Burns \etal~\cite{burns2023weak}, our work pivots towards a scenario where the student network surpasses the teacher in visual capabilities. Despite this inversion of roles, there remains valuable dark knowledge in the teacher network that can be transferred to the student, either through logits or via intermediate representational features. To benchmark our experiments, we employ a range of established~\cite{kd,hint,rkd,ofd,revkd,vanillakd} and recently proposed~\cite{dkd,burns2023weak} distillation techniques as baseline methods.

\subsection{Main Results}

\subsubsection{Image Classification.}
\textbf{CIFAR-100 image classification.} We commence our investigation with an exploration of weak-to-strong generalization (WSG) on the CIFAR-100 dataset. The outcomes of this investigation are delineated in Tables~\ref{tab:cifar_cls_same} and ~\ref{tab:cifar_cls_different}. Specifically, Table~\ref{tab:cifar_cls_same} presents the scenarios in which both teacher and student models share the same network architectures. We examine a range of prevalent vision architectures such as ResNet~\cite{resnet}, WRN~\cite{wrn}, and VGG~\cite{vgg}. We employ various KD methods to assess the potential of larger-capacity students guided by limited-capacity teachers. Remarkably, in nearly all cases employing KD-based approaches, the student models outperform those trained from scratch.
Furthermore, both AugConf~\cite{burns2023weak} and our proposed AdaptConf method surpasses all previous distillation techniques across all teacher-student pairs. This highlights that simply emulating a weak teacher does not yield the most favorable outcomes. Notably, AdaptConf consistently achieves superior performance compared to AugConf~\cite{burns2023weak}, underscoring the advantage of our dynamic adaptive confidence weighting. This approach provides a more refined mechanism for facilitating weak-to-strong knowledge transfer.

\begin{table}[t]
\renewcommand{\arraystretch}{1.05}
\setlength{\tabcolsep}{5pt}
\setlength{\belowcaptionskip}{-10pt}
\center
\begin{small}
\begin{tabular}{c|c|c}
\multirow{2}{*}{Teacher} & ResNet18 & MobileNet-V1 \\
& 69.75 & 71.57 \\
\multirow{2}{*}{Student} & ResNet34 & ResNet50 \\
& 73.47 & 76.22 \\
\Xhline{3\arrayrulewidth} 
KD~\cite{kd} & 73.68 & 76.52 \\
FitNet~\cite{hint} & 70.93 & 73.61 \\
RKD~\cite{rkd} & 73.65 & 76.45 \\
ReviewKD~\cite{revkd} & 72.99 & 75.28 \\
DKD~\cite{dkd} & 73.74 & \underline{76.72} \\
AugConf~\cite{burns2023weak} & \underline{73.80} & 76.64\\
\hline
AdaptConf~(\textbf{Ours}) & \textbf{74.16} & \textbf{76.94} \\
$\Delta$ & \textcolor{forestgreen}{+0.69} & \textcolor{forestgreen}{+0.72} \\
\end{tabular}
\end{small}
\vspace{-4pt}
\caption{\textbf{Top-1 results on the ImageNet validation set.} $\Delta$ represents the performance improvement over the student model trained from scratch.}
\label{tab:imagenet_cls}
\end{table}

Table~\ref{tab:cifar_cls_different} presents the results for teacher-student pairs from different series, such as ShuffleNet~\cite{shufflenet} and MobileNet~\cite{mobilenetv2}. Additionally, take the MobileNetV2-ResNet50 pair as an example, the experimental results reveal that when the teacher model is significantly weaker, \ie a substantial performance gap exists between the weak teacher model and the strong student model, none of the KD-based methods were able to effectively enhance the strong student's performance, except for AugConf and AdaptConf. The possible reason is that these methods include the predictions of the strong student in the loss function. This proves that self-training methods, akin to those described in ~\cite{lee2013pseudo}, can mitigate the bias from a suboptimal teacher model.
It is important to note that FitNet~\cite{hint} consistently underperforms when compared to training from scratch. This could be attributed to its sole focus on intermediate features, which may be more misleading for the strong student to learn from than soft predictions, as suggested by ~\cite{ofakd}. Overall, our AdaptConf achieves an improvement of 0.5\%-2\% on all evaluated teacher-student pairings, whether they are from the same or different series.

Furthermore, we investigate a scenario where only the teacher's output is available, as shown in Table~\ref{tab:cifar_cls_different_no_gt}. In this context, it becomes evident that AugConf and AdaptConf yields more significant improvements compared to other KD-based methods when ground truth is absent. This observation underscores the suitability of our confidence distillation approach for more extreme WSG scenarios where ground truth is not available.

\textbf{ImageNet image classification.}
Table~\ref{tab:imagenet_cls} presents the top-1 accuracy for image classification on the ImageNet dataset. Our AdaptConf method achieves significant improvements across both WSG scenarios, whether employing the same or different architectures.

\begin{table}[!ht]
    \small
    \renewcommand{\arraystretch}{1.05}
    \setlength{\tabcolsep}{4pt}
    \setlength{\belowcaptionskip}{-10pt}
    \center
    \begin{tabular}{c|cccc}
        \multirow{2}{*}{Teacher} & \multicolumn{2}{c}{ResNet12} & \multicolumn{2}{c}{ResNet18} \\
        ~                            & 59.65                          & 77.80                          & 60.83                          & 78.96                          \\
        \multirow{2}{*}{Student} & \multicolumn{2}{c}{ResNet36} & \multicolumn{2}{c}{ResNet36} \\
        ~                            & 60.91                          & 79.01                          & 60.91                          & 79.01                          \\
        \Xhline{3\arrayrulewidth}
        ~                            & 1-shot                         & 5-shot                         & 1-shot                         & 5-shot                         \\
        \Xhline{3\arrayrulewidth}
        KD                           & 60.94                          & 79.14                          & 61.57                          & \underline{79.79}              \\
        RKD~\cite{rkd}               & 59.74                          & 78.30                          & 60.80                          & 78.82                          \\
        AugConf~\cite{burns2023weak} & \underline{61.38}              & \underline{79.33}              & \underline{61.66}              & 79.46                          \\
        \hline
        AdaptConf~(\textbf{Ours})    & \textbf{61.50}                 & \textbf{79.52}                 & \textbf{62.29}                 & \textbf{79.96}                 \\
        $\Delta$                     & \textcolor{forestgreen}{+2.59} & \textcolor{forestgreen}{+2.67} & \textcolor{forestgreen}{+3.38} & \textcolor{forestgreen}{+3.11}    \\
    \end{tabular}
    \caption{\textbf{Average 5-way accuracy (\%) with 95\% confidence interval on the miniImageNet validation set in Classification Training stage.} $\Delta$ represents the performance improvement over the student model trained from scratch. All results are the average over 3 trials.}
    \label{tab:meta_base-class}
\end{table}

\begin{table*}[!ht]
    \renewcommand{\arraystretch}{1.05}
    \setlength{\tabcolsep}{5pt}
    \setlength{\belowcaptionskip}{-10pt}
    \center
    \begin{small}
        \begin{tabular}{c|cc|cc}
            \multirow{2}{*}{Teacher}     & ResNet12 (Class-stage)          & ResNet18 (Class-stage)          & ResNet12 (Meta-stage)          & ResNet18 (Meta-stage)            \\
                                        & 59.20                          & 60.63                          & 65.26                          & 66.51                               \\
            \multirow{2}{*}{Student}     & ResNet36                       & ResNet36                       & ResNet36                       & ResNet36                       \\
                                        & 65.08                          & 65.08                          & 65.08                          & 65.08                          \\
            \Xhline{3\arrayrulewidth}
            KD~\cite{kd}                 & 63.43                          & 65.04                          & \underline{66.08}              & \underline{65.93}              \\
            RKD~\cite{rkd}               & 64.79                          & 65.42                          & 65.96                          & 65.46                          \\
            AugConf~\cite{burns2023weak} & \underline{65.15}              & \underline{65.59}              & 65.9                           & 65.78                          \\
            \hline
            AdaptConf~(\textbf{Ours})    & \textbf{65.38}                 & \textbf{65.74}                 & \textbf{66.08}                 & \textbf{65.95}                 \\
            $\Delta$                     & \textcolor{forestgreen}{+0.30} & \textcolor{forestgreen}{+0.66} & \textcolor{forestgreen}{+1.00} & \textcolor{forestgreen}{+0.87}    \\
        \end{tabular}
    \end{small}
    \vspace{-1pt}
    \caption{\textbf{Average 5-way accuracy on the miniImageNet validation set at Meta-Learning stage.} $\Delta$ represents the performance improvement over the student model trained from scratch. All results are the average over 3 trials.}
    \label{tab:meta_base-meta}
\end{table*}

\begin{table*}
    \renewcommand{\arraystretch}{1.05}
    \setlength{\tabcolsep}{10pt}
    \setlength{\belowcaptionskip}{-10pt}
    \small
    \center
    \begin{tabular}{c|cccc|cccc}
        dataset                      & \multicolumn{4}{c|}{CIFAR-10}                                                                                                      & \multicolumn{4}{c}{CIFAR-100}                                                                                                     \\
        \hline
        noise type                   & \multicolumn{2}{c}{asymmetric}                                  & \multicolumn{2}{c|}{symmetric}                                   & \multicolumn{2}{c}{asymmetric}                                  & \multicolumn{2}{c}{symmetric}                                   \\
        \hline
        \multirow{2}{*}{Teacher}     & \multicolumn{2}{c}{PR18}                                        & \multicolumn{2}{c|}{PR18}                                        & \multicolumn{2}{c}{PR18}                                        & \multicolumn{2}{c}{PR18}                                        \\
                                     & 92.98                          & 99.56                          & 95.80                           & 99.80                           & 73.20                           & 92.67                          & 76.16                          & 92.90                           \\
        \multirow{2}{*}{Student}     & \multicolumn{2}{c}{PR34}                                        & \multicolumn{2}{c|}{PR34}                                        & \multicolumn{2}{c}{PR34}                                        & \multicolumn{2}{c}{PR34}                                        \\
                                     & 93.69                          & 99.61                          & 96.13                          & 99.77                          & 74.80                           & 92.94                          & 78.20                           & 93.77                          \\
        \Xhline{3\arrayrulewidth}
                                     & Top-1                           & Top-5                           & Top-1                           & Top-5                           & Top-1                           & Top-5                           & Top-1                           & Top-5                           \\
        \Xhline{3\arrayrulewidth}
        KD~\cite{kd}                 & \underline{93.54}              & \underline{99.84}              & 95.90                           & 99.84                          & \underline{75.49}              & 93.67                          & 77.61                          & 93.74                          \\
        RKD~\cite{rkd}               & 92.42                          & 99.75                          & \underline{95.99}              & \underline{99.85}              & 74.20                           & 93.54                          & 76.92                          & 93.09                          \\
        AugConf~\cite{burns2023weak} & 92.60                           & 99.75                          & 95.10                           & 99.83                          & 74.99                          & \underline{93.72}              & \underline{78.34}              & \underline{94.02}              \\
        \hline
        AdaptConf~(\textbf{Ours})    & \textbf{93.69}                 & \textbf{99.84}                 & \textbf{96.13}                 & \textbf{99.87}                 & \textbf{75.61}                 & \textbf{93.78}                 & \textbf{78.64}                 & \textbf{94.03}                 \\
        $\Delta$                     & \textcolor{forestgreen}{+0.00} & \textcolor{forestgreen}{+0.23} & \textcolor{forestgreen}{+0.00} & \textcolor{forestgreen}{+0.10} & \textcolor{forestgreen}{+0.81} & \textcolor{forestgreen}{+0.84} & \textcolor{forestgreen}{+0.44} & \textcolor{forestgreen}{+0.26} \\
    \end{tabular}
    \vspace{-4pt}
    \caption{\textbf{Top-1 and top-5 results on the CIFAR-10/CIFAR-100 noise label validation set.} $\Delta$ represents the performance improvement over the student model trained from scratch. All results are the average over 3 trials.}
    \label{tab:lnl}
\end{table*}

\subsubsection{Few-shot learning}
For the few-shot learning task, we conduct distillation experiments separately in the classification (Table~\ref{tab:meta_base-class}) and meta-learning (Table~\ref{tab:meta_base-meta}) stages. We compare and evaluate the performances of student when trained with teachers of different sizes. 
In the classification experiments, only RKD results in a performance degradation of the student model, while the usage of other methods led to varying degrees of improvement. Notably, our confidence-based method outperforms previous knowledge distillation based ones.
In the meta-learning stage, we employ weights from different training stages of the same model as the teacher. Experimental results demonstrate significant advantages of our proposed method. Even when using the Class-stage weight as the teacher, our approach achieves a +0.66\% improvement over the baseline set by a weaker ResNet18(Class-stage) teacher model.  Furthermore, when using the same stage weight as the teacher, our confidence-based method surpasses previous knowledge distillation results to a greater extent.

\subsubsection{Transfer learning}
Table~\ref{tab:transfer} examines the efficacy of transfer learning using the iNaturalist~\cite{inat} and ImageNet~\cite{deng2009imagenet} datasets. When our method is trained with ground truth labels on ImageNet, it demonstrates a notable enhancement, achieving an increase of +0.33\% in top-1 accuracy on a model with a high precision of 83.5\%. Even without ground truth labels, our approach still secures a +2.15\% improvement over the baseline set by a weaker ResNet50 teacher model. On the iNaturalist dataset, our confidence-based method also surpasses previous knowledge distillation results by a considerable margin.

\subsubsection{Learning with noisy labels}
In Table~\ref{tab:lnl}, we analyze the effectiveness of weak-to-strong using the CIFAR-10 and CIFAR-100 datasets under two simulated noisy label settings. When training the model on the sample dataset (CIFAR-10), all methods except ours, negatively impact the model given its already high accuracy. This underscores the robustness of our method, irrespective of the performance gap between the teacher and student models.
On the CIFAR-100 dataset, our method demonstrates a performance improvement of 0.81\% in top-1 accuracy under the asymmetric noise type setting.

\begin{figure*}[t]
\centering
\subfloat{\includegraphics[height=38mm]{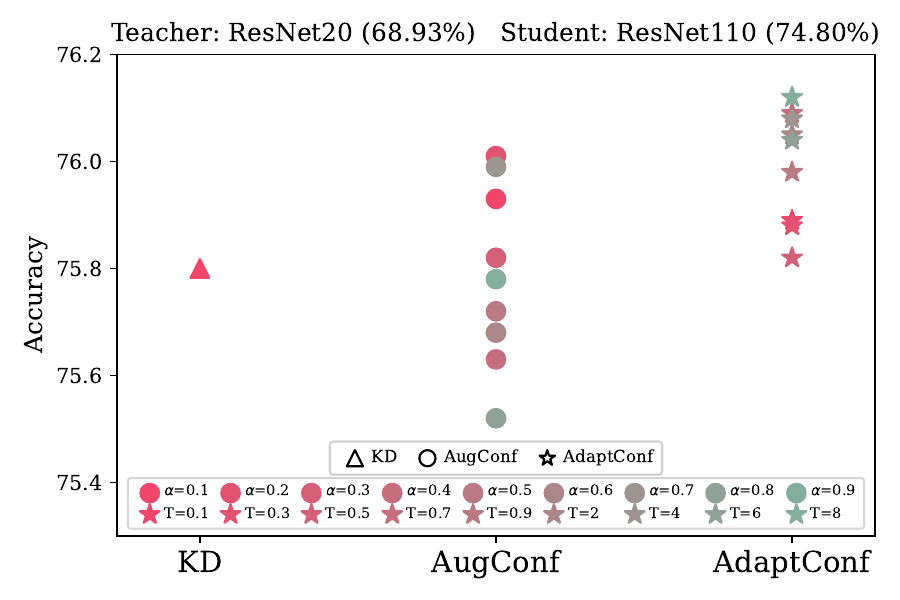}}
\subfloat{\includegraphics[height=38mm]{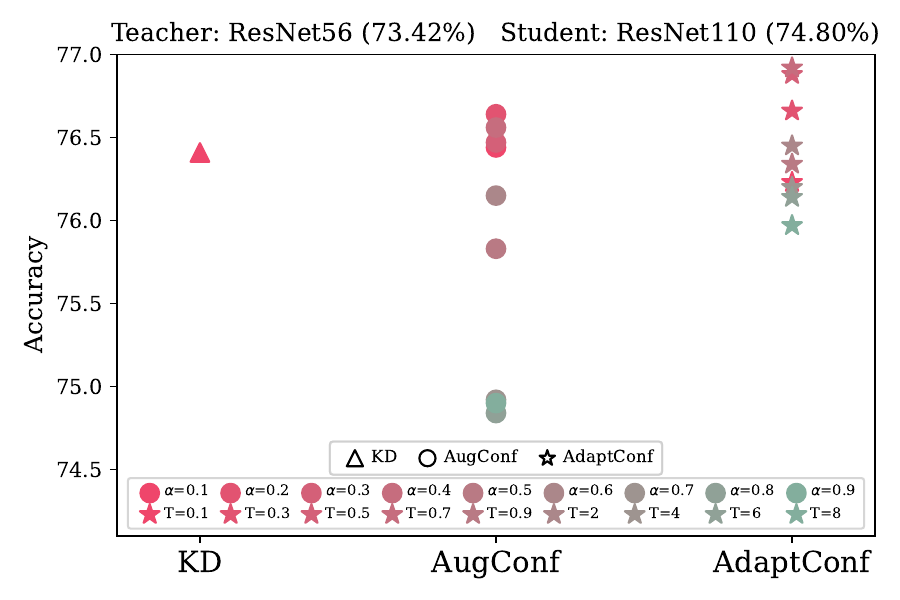}} 
\subfloat{\includegraphics[height=38mm]{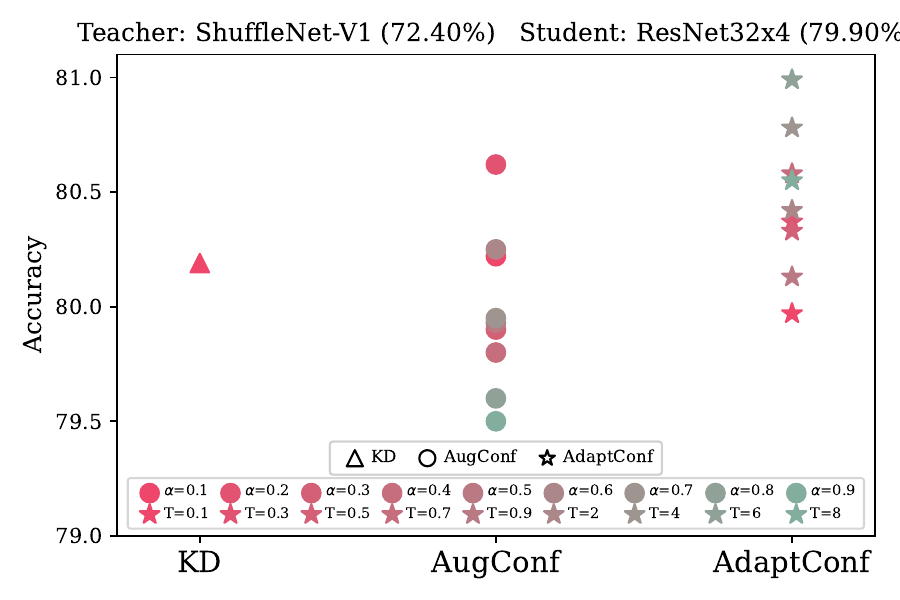}}
\vspace{-1pt}
\caption{\small{Ablation study examining the impact of hyper-parameter variation on confidence distillation results. The parameter $\alpha$ for AugConf is adjusted across a range from 0.1 to 0.9, while the temperature $T$ for AdaptConf is scaled from 0.1 to 8.}}
\label{fig:abla_hyper}
\end{figure*}

\begin{figure*}[t]
\centering
\subfloat{\includegraphics[width=1\linewidth]{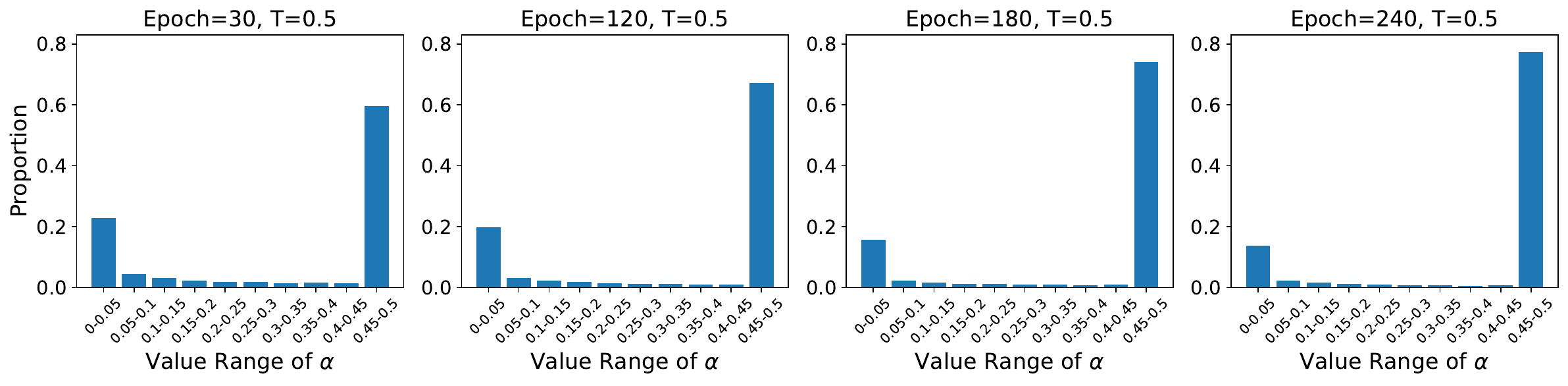}}
\\
\subfloat{\includegraphics[width=1\linewidth]{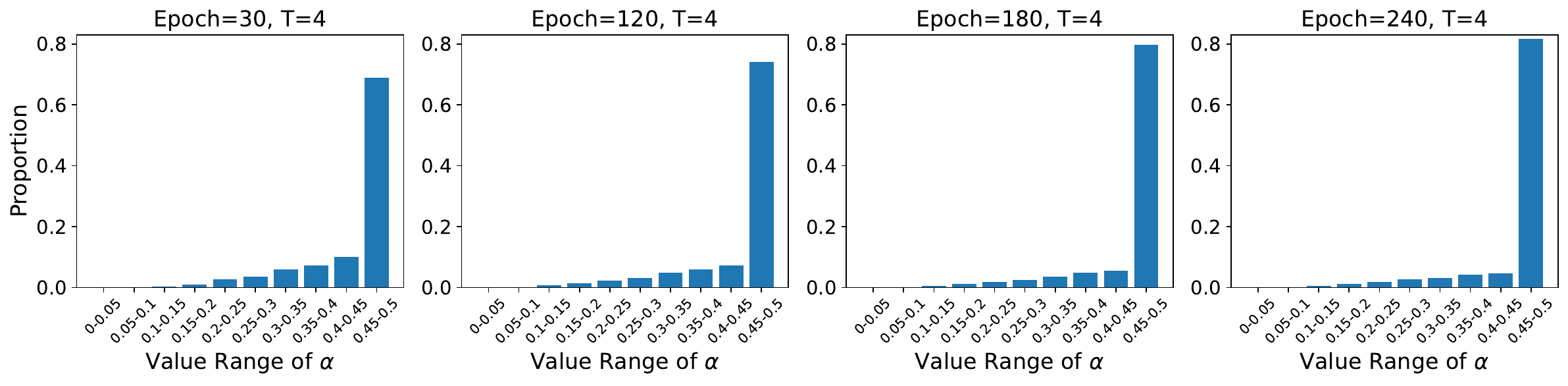}}
\vspace{-1pt}
\caption{\small{Quantitative analysis about the value of $\beta$(x) in Eq.~\ref{eq:adaptconf} on the CIFAR-100 dataset. The evaluation is based on the ShuffleNetV1-ResNet32x4 teacher-student architecture pair.}}
\label{fig:abla_balance}
\end{figure*}

\subsection{Ablation Study}

\textbf{Robustness of confidence distillation.}
In this study, we investigate the necessity of devising a method that goes beyond a mere weighted combination of labels. As depicted in Eq.~\ref{eq:augconf}, despite its straightforward approach of integrating direct learning from a weaker model with the intrinsic capacity of a stronger model, AugConf~\cite{burns2023weak} still requires manual tuning of a hyper-parameter $\alpha$ to balance the ratio of two different objectives. The setting of different $\alpha$ values can have varying impacts across different contexts. Similarly, although our proposed AdaptConf does not require manual adjustment of $\alpha$ to balance the proportions of objectives, we can manipulate the temperature $T$ to control the degree of probability distribution in soft labels during the computation of the cross-entropy CE($\cdot$), following a conventional distillation method~\cite{kd}. Therefore, we explore the effects of these two methods under different hyper-parameter settings on the final outcome.
Overall, the performance of KD, AugConf, and AdaptConf improves sequentially across various architectural settings. Moreover, it can be observed that AugConf exhibits a larger fluctuation in results compared to AdaptConf, indicating that the influence of $\alpha$ on AugConf is more significant than the effect of $T$ on AdaptConf, which suggests that our AdaptConf has superior robustness. Additionally, the average outcomes achieved by AdaptConf are consistently higher than those of AugConf under different hyper-parameter settings.

\textbf{Robustness of confidence distillation.}
In this section, we perform a quantitative analysis of the confidence weight determined by our dynamic function $\beta(x)$ as delineated in Eq.~\ref{eq:adaptconf}, with the findings illustrated in Figure~\ref{fig:abla_balance}. We selected checkpoints from four distinct training phases and calculated their specific $\beta(x)$ values on the validation set. It can be observed that as training progresses, the proportion of samples with $\beta = 0.5$ increases, indicating that the student model's performance is improving and being aligned with the weak teacher's correct classifications. A higher temperature setting $T$ reduces the cross-entropy (CE) discrepancy between the teacher and student, promoting a more uniform balance between the weak teacher's guidance and the strong student's own predictions. Consequently, the number of samples where $\beta = 0.5$ also increases with training. These phenomena collectively validate that our proposed AdaptConf can dynamically adjust the learning ratio between the two components.

\section{Conclusion}

In this paper, we investigate weak-to-strong generalization for vision foundation models and unveil a promising avenue for enhancing the capabilities of artificial intelligence in the visual domain. By leveraging an innovative adaptive confidence loss mechanism, we demonstrate the feasibility and effectiveness of using weaker models to supervise and improve stronger counterparts. Our findings not only validate the potential of weak-to-strong generalization but also set the stage for future research endeavors aimed at unlocking further advancements in AI model performance. This work contributes a significant step forward in the pursuit of more sophisticated, efficient, and capable AI systems, emphasizing the importance of nuanced supervision mechanisms in achieving superhuman performance in vision tasks.


\bibliography{egbib}
\bibliographystyle{icml2024}

\newpage
\appendix
\onecolumn
\section{Implementation details.}

\subsection{ImageNet Classification}
\textbf{CIFAR-100.}
We adopt the vision architectures of the teacher and student models as outlined in the traditional distillation papers~\cite{ofakd,dkd}. It should be noted that the previous codebase~\cite{dkd} conducted experiments on CIFAR-100 using only 1 GPU. To expedite our experiments, we leverage the \emph{distributed} Pytorch framework~\cite{pytorch} to train and do inference on 8 GPUs. Consequently, some hyperparameter settings and results may not align exactly with the previous paper. Specifically, we employ the SGD optimizer with a momentum of 0.9. The learning rate starts at 0.2 and decays with a minimum learning rate of 2e-3 using a cosine annealing schedule. We train for 240 epochs with a batch size of 512 spread across 8 GPUs, and apply a weight decay of 0.0005. Standard data augmentation techniques, including random resized crop and horizontal flip are utilized.

\textbf{ImageNet.}
we employ the SGD optimizer with a momentum of 0.9. The learning rate starts at 0.1 and decays with a rate of 0.1 every 30 epochs. We train for 100 epochs with a batch size of 512 spread across 8 GPUs, and apply a weight decay of 0.0001. Standard data augmentation techniques, including random resized crop, horizontal flip and label smoothing are utilized.

\subsection{Transfer learning.}
To fine-tune the self-supervised pretrained ViT-B on ImageNet and iNaturalist, we adopt the hyperparameter settings from MAE~\cite{mae}. The adamw optimizer is employed for this purpose. The learning rate begins at 2e-3 and gradually decays with a minimum learning rate of 1e-6, utilizing a cosine annealing schedule. We conduct training for 100 epochs, utilizing a batch size of 4096 across 8 GPUs. A weight decay of 0.05 is applied to mitigate overfitting. The fine-tuning process incorporates robust data augmentation techniques, including auto-augment, mixup, cutmix, and stochastic drop path.

\subsection{Few-shot Leaerning}
We use ResNet12 and follow the setting of~\cite{chen2021meta} on miniImageNet dataset, and created ResNet18 and ResNet36 by increasing the number of layers in original ResNet12. For the classification training stage, we use the SGD optimizer with momentum 0.9. The learning rate starts from 0.1 and the decay factor is set to 0.1. On miniImageNet, we train 100 epochs with the batch size of 128 on 4 GPUs, the learning rate decays at 90 epoch, and the weight decay is 0.0005. Standard data augmentation strategies including random resized crop and horizontal flip are applied. For meta-learning stage, we use the SGD optimizer with momentum 0.9. The learning rate is fixed as 0.001. The batch size is set to 4, \ie, each training batch contains 4 few-shot tasks to compute the average loss.  The cosine scaling parameter $\tau$ is initialized as 10.
For knowledge distillation, the kd loss weight is set to 1, the temperature is set to 10. We use the threshold with 8 and 0.25 for classifier stage and meta stage, respectively.

\subsection{Learning with noisy labels}
For CIFAR-10/100 datasets, we follow~\cite{li2022selective} use a PreAct ResNet18 network, and created PreAct ResNet34 by increasing the number of layers in PreAct ResNet12.
We train our models using SGD with a momentum of 0.9, a weight decay of 1e−4, and a batch size of 128. The network is trained for 250 epochs and the warm-up epoch is set to 1 dufring training stage. We set the initial learning rate as 0.1, and reduce it by a factor of 10 after 125 and 200 epochs. The fine-tuning stage of Sel-CL+ has 70 epochs, where the learning rate is 0.001. We always set the Mixup hyperparameter to 1, scalar temperature to 0.1, and loss weights to 1. We try two settings of simulated noisy labels: symmetric and asymmetric. And the noise ratio is set to 0.2 and 0.4, respectively.
For knowledge distillation, we set the threshold to 0.5 and assign a weight of 1 to the knowledge distillation loss.


\end{document}